\documentclass[conference]{IEEEtran}
\usepackage{lineno,hyperref}
\usepackage{multirow}
\usepackage{graphicx}
\usepackage{amsmath}
\usepackage{amssymb}

\ifCLASSINFOpdf
 
\begin{document}

\title{Parameter estimation of the homodyned K distribution based on neural networks and trainable fractional-order moments}

\author{\IEEEauthorblockN{Michal Byra\IEEEauthorrefmark{1}\IEEEauthorrefmark{2}, Ziemowit Klimonda\IEEEauthorrefmark{2}, Piotr Jarosik\IEEEauthorrefmark{2}}\\

\IEEEauthorblockA{\IEEEauthorrefmark{2}Institute of Fundamental Technological Research, \\Polish Academy of Sciences, Warsaw, Poland}

\IEEEauthorblockA{\IEEEauthorrefmark{1}Corresponding author, e-mail: mbyra@ippt.pan.pl}
}

\maketitle

\begin{abstract}

Homodyned K (HK) distribution has been widely used to describe the scattering phenomena arising in various research fields, such as ultrasound imaging or optics. In this work, we propose a machine learning based approach to the estimation of the HK distribution parameters. We develop neural networks that can estimate the HK distribution parameters based on the signal-to-noise ratio, skewness and kurtosis calculated using fractional-order moments. Compared to the previous approaches, we consider the orders of the moments as trainable variables that can be optimized along with the network weights using the back-propagation algorithm. Networks are trained based on samples generated from the HK distribution. Obtained results demonstrate that the proposed method can be used to accurately estimate the HK distribution parameters.

\end{abstract}


\IEEEpeerreviewmaketitle

\section{Introduction}

Homodyned K (HK) distribution has been widely used to describe the scattering phenomena arising in various research fields. In ultrasound (US) imaging, the HK distribution has been utilized to model the backscattered echo amplitude and quantitatively assess tissue structure \cite{oelze2016review}. For example, the HK distribution was applied for ultrasound based temperature monitoring and tissue characterization \cite{byra2016classification,byra2017temperature,tsai2021ultrasound,roy2018assessment}. 

Various methods have been developed for the estimation of the HK distribution parameters. Hruska and Oelze proposed a level-set estimation technique based on the signal-to-noise ratio, skewness and kurtosis parameters calculated using fractional-order moments \cite{hruska2009improved}. Destrempes et al. proposed an iterative estimation technique based on the first moment of the intensity and two log-moments, namely the $X$- and $U$-statistics \cite{destrempes2013estimation}. Building on the previous works, Zhou et al. utilized an artificial neural network (ANN) to estimate the parameters of the HK distribution \cite{zhou2021parameter}. Authors utilized the signal-to-noise ratio, skewness, kurtosis, $X$- and $U$- statistics as the input to the feed-forward neural network. 

In this work, we propose a machine learning based technique for the estimation of the HK distribution parameters. Similar to Zhou et al., we train our neural network based on the SNR, skewness and kurtosis statistics \cite{zhou2021parameter}. However, in our case the orders of the moments used for the calculations are not fixed. Hruska and Oelze presented that the choice of the moments is important for the accurate estimation of the HK distribution parameters \cite{hruska2009improved}. To improve the estimation, we treat the orders of the moments as trainable variables that can be optimized along with the network weights using the back-propagation algorithm.  

\begin{figure*}[]
	\begin{center}
		\includegraphics[width=0.9\linewidth]{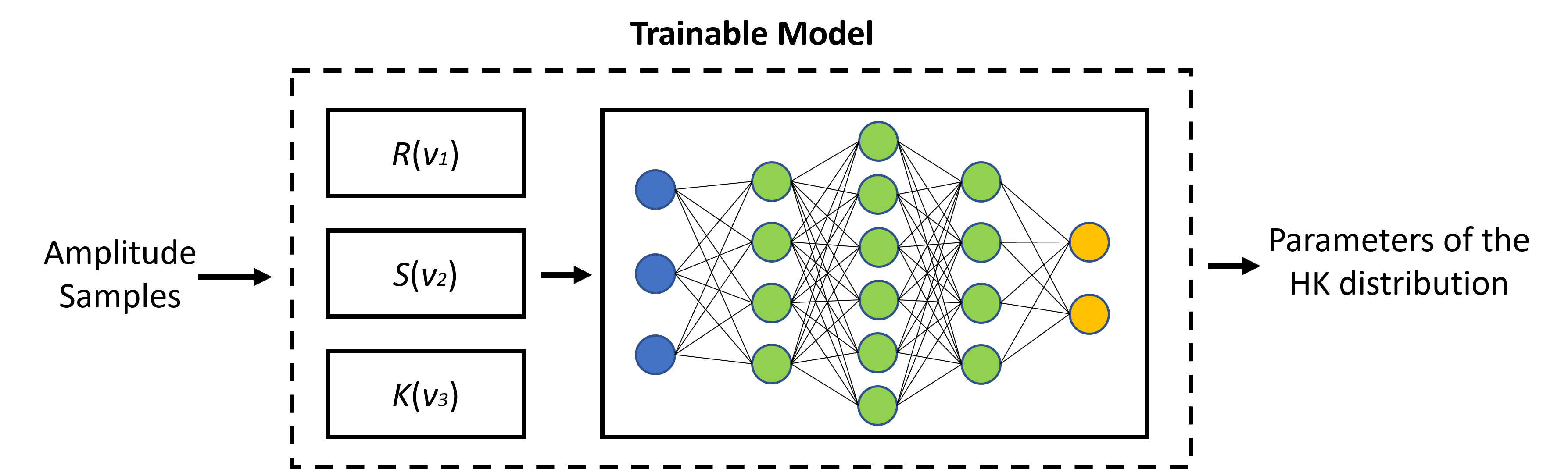}
	\end{center}
	\caption{Scheme presenting the proposed approach to the estimation of the HK distribution parameters. In our work, the order of the moments, related to the $v$ value, used to calculate SNR $R$, skewness $S$ and kurtosis $K$ are trainable in the same way as the weights of the neural network. }
	\label{f1}
\end{figure*}

\section{Methods}

\subsection{Homodyned K distribution}

The probability density function of the HK distribution can be expressed in the following way: 

\begin{equation} 
p(A) = A \int\limits_{0}^{\infty} hJ_{0}(sh)J_{0}(Ah) \Big(1 + \dfrac{h^2 \sigma^2}{2u} \Big)^{-u}dh,
\end{equation}

\noindent where $A$ stands for the amplitude, $J_{0}$ is the zero-th order Bessel function of the first kind and variable $h$ is used for the integration. Parameters $s^2$ and $\sigma^2$ stand for the coherent and diffusive signal power. HK distribution has two parameters used for the quantitative assessment of the scattering phenomena in US. The first parameter, $u$, is the scatterer clustering parameter reflecting the number of the scatterers in the resolution cell. The second quantitative parameter of the HK distribution is expressed as the ratio $k=\frac{s}{\sigma}$ and is related to the spatial periodicity of the scatterer distribution.

\subsection{The RSK estimator}

Hruska and Oelze proposed the level-set method for the estimation of the HK distribution parameters based on the signal-to-noise ratio (R), skewness (S) and kurtosis (K) of the amplitude, denoted as the RSK estimator \cite{hruska2009improved}. These three can be calculated with the following equations: 

\begin{equation} 
R(v) = \dfrac{\text{E}[A^v]}{(\text{E}[A^{2v}] - \text{E}^2[A^v] )^{1/2}},
\end{equation}

\begin{equation} 
S(v) = \dfrac{\text{E}[A^{3v}] - 3\text{E}[A^{v}]\text{E}[A^{2v}] + 2\text{E}^3[A^{v}]}{(\text{E}[A^{2v}] - \text{E}^2[A^v] )^{3/2}},
\end{equation}

\begin{equation} 
K(v) = \dfrac{\text{E}[A^{4v}] - 4\text{E}[A^{v}]\text{E}[A^{3v}] + 6\text{E}[A^{2v}]\text{E}^2[A^{v}] - 3\text{E}^4[A^{v}]}{(\text{E}[A^{2v}] - \text{E}^2[A^v] )^{2}},
\end{equation}

\begin{figure*}[]
	\begin{center}
		\includegraphics[width=0.9\linewidth]{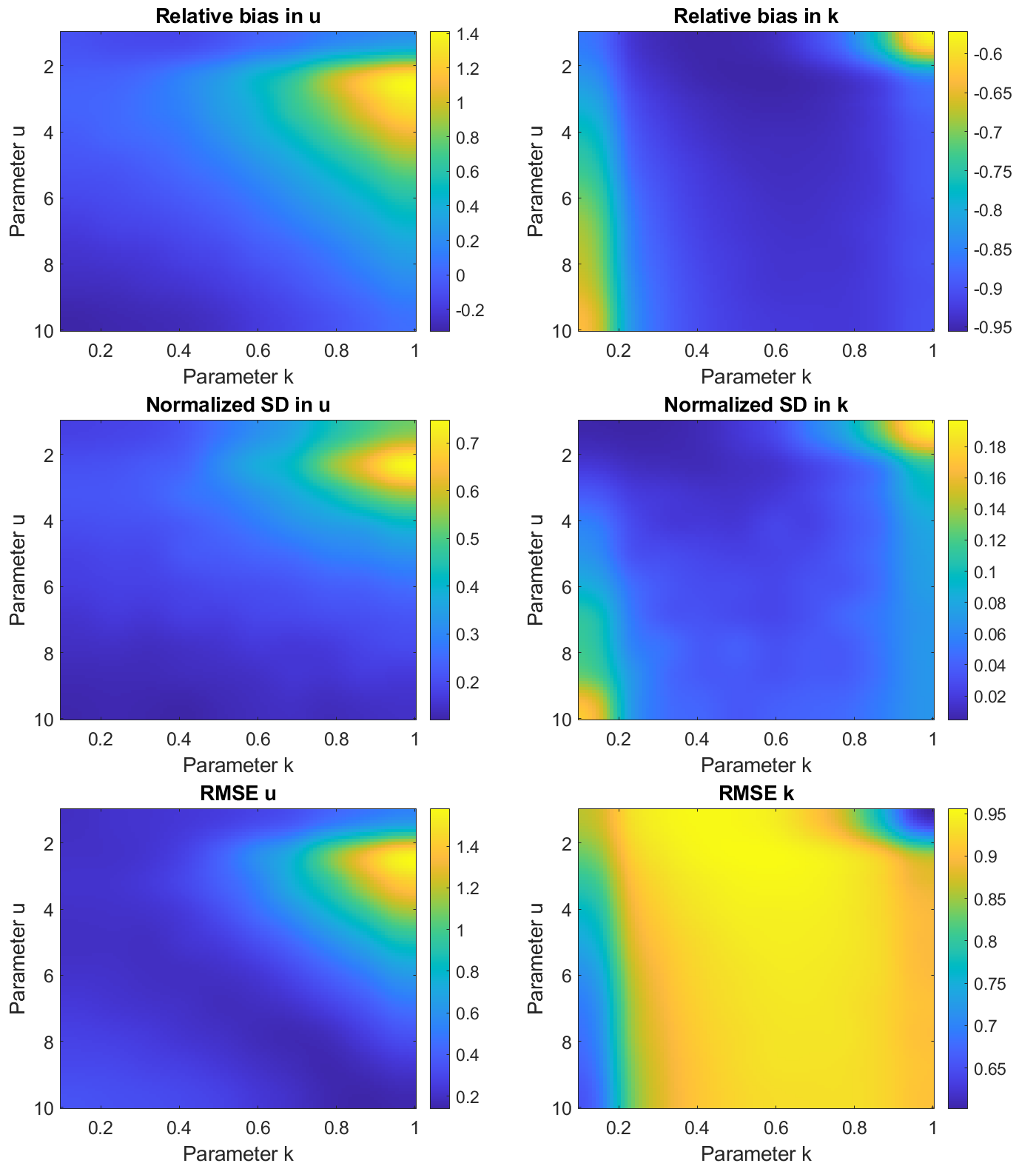}
	\end{center}
	\caption{Relative biases, normalized standard deviations (SD) and rooted mean squared errors obtained for the HK distribution parameters estimator based on the ensemble of neural networks. }
	\label{f2}
\end{figure*}

\begin{figure}[]
	\begin{center}
		\includegraphics[width=0.7\linewidth]{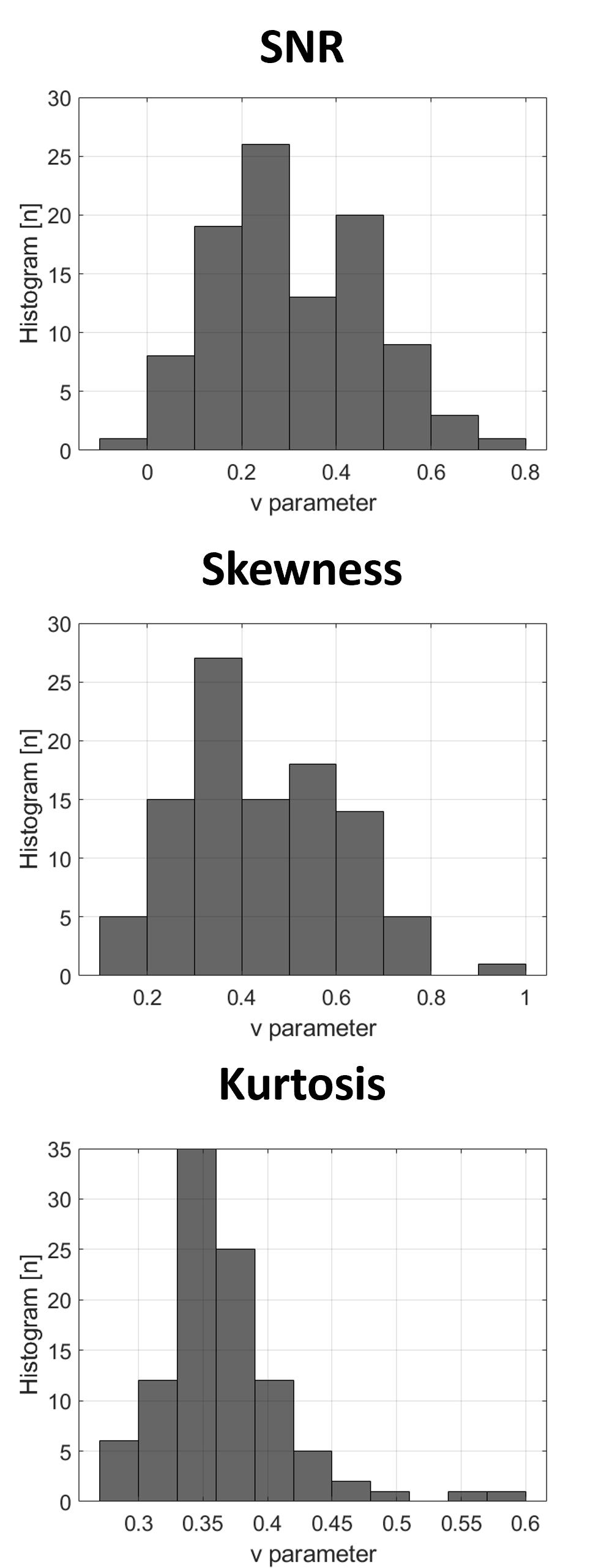}
	\end{center}
	\caption{Histograms of the $v$ values obtained in the training procedure for the SNR $R$, skewness $S$ and kurtosis $K$ for networks constituting the ensemble. }
	\label{f3}
\end{figure}

\noindent where $v$ is a positive number used to adjust the orders of the amplitude moments. In the case of the level-set approach, the $R$, $S$ and $K$ calculated based on the amplitude samples are compared with the theoretical values of these parameters determined for the HK distribution with specific $u$ and $k$ parameters. Hruska and Oelze presented that the choice of the $v$ parameter in eq. 2-4 has a large impact on the performance of the level-set method \cite{hruska2009improved}. Authors reported that the better estimation performance of the level-set method could be achieved based on six level curves corresponding to $R$, $S$ and $K$ calculated for two values of the $v$ parameter, namely 0.72 and 0.88. These values of the $v$ parameter were selected using the grid-search algorithm for $v$ ranging from 0.02 to 1, with an increment of 0.02. 

\subsection{Neural network based estimation}

Zhou et al. developed a neural network to estimate the parameters of the HK distribution based on the $R$, $S$, $K$, $X$- and $U$-statistics \cite{zhou2021parameter}. $R$, $S$ and $K$ were calculated for a fixed value of the $v$ parameter equal to 2. Similar to Zhou et al., we utilize a feed-forward neural network to determine the $u$ and $k$ parameters of the HK distribution. However, in our case we treat the $v$ parameter as a trainable variable that can be separately optimized for R, S and K in eq. 2-4. During the training of the network, we utilize the back-propagation algorithm to adjust the values of the $v$ parameters. Compared to the grid-search algorithm used by Hruska and Oelze to select $v$, in our work the order of the moments are determined automatically \cite{hruska2009improved}.  Scheme of the proposed method is presented in Fig. \ref{f1}. The input of the network consisted of three units corresponding to the $R$, $S$ and $K$ parameters. Next, three hidden dense layers were utilized with the number of units set to 6, 12 and 6, respectively. The output of the network consisted of 2 units designed to calculate the $u$ and $k$ parameters. Additionally, each dense layer was equipped with the batch-normalization layer and sigmoid activation function. Following the work of Zhou et al., the network was trained to output the $u$ and $k$ parameters for ranges of $log_{10}(u)\in [-1, 2] $ and $k \in [0, 2]$, additionally taking into account the scaling required for the sigmoid activation function \cite{zhou2021parameter}. The mean absolute percentage error (MAPE) and the Adam optimizer with the learning rate of 0.001 were used for the training. Network was trained based on samples generated from the HK distribution for $log_{10}(u)\in [-1, 2] $ and $k \in [0, 2]$. Batch size was set to 16. For each batch, the number of the amplitude samples was drawn at random from the interval [500, 2000]. TensorFlow was utilized for the calculations \cite{abadi2016tensorflow}.

\subsection{Evaluation}

We followed the same approach to the evaluation as in the previous studies \cite{hruska2009improved,destrempes2013estimation}. Amplitudes were sampled from the HK distribution with $u$ and $k$ parameters in the domains of $u \in \{1, 2, ..., 9, 10 \}$ and $k \in \{0.1, 0.2, ..., 0.9, 1.0 \}$, respectively. The mean intensity of the HK distribution was constant. For each pair of the parameters, we simulated 1000 sets, each with the number of samples equal to 1000. Based on the estimates $\hat{u}$ and $\hat{k}$ determined for each set, we calculated the relative biases $(\text{E}[\hat{u}] - u) /u$ and $(\text{E}[\hat{k}] - k) /k$ as well as the normalized standard deviations $\sqrt{\text{Var}[\hat{u}]}/u$ and $\sqrt{\text{Var}[\hat{k}]}/k$. Moreover, the relative root mean squared errors (RMSEs) $\sqrt{(\text{E}[\hat{u}] - u)^2} /u$ and $\sqrt{(\text{E}[\hat{k}] - k)^2} /k$ were computed. The proposed approach was evaluated in two settings. First, we assessed the estimation performance of an ensemble including 100 networks trained with different initial weights. In this case, the outputs of the networks were averaged to obtain the final prediction. Second, the better performing network in respect to the average RMSEs was selected and evaluated separately. The initial values of the $v$ parameters were set to 0.5, which corresponded to the mid-point of the range used by Hruska and Oelze in the case of the grid-search range of [0, 1]  \cite{hruska2009improved}.  Additionally, the proposed approach was compared with the RSK 
 and XU estimators as well as with a network trained with the $v$ value equal to 2 as in Zhou et al. \cite{zhou2021parameter}. Evaluations were performed in Matlab (MathWorks, USA).

\section{Results and discussion}

Table \ref{t1} presents the performance of the network ensemble. Here, we can observe that the method based on networks with trainable orders of moments achieved better performance then the network with the fixed values of $v$ equal to 2. Similar results are presented in Table \ref{t2} for the single networks. Moreover, both Tables present that the network based approaches outperformed the RSK and XU estimators for the majority of metrics. Error plots calculated for the ensemble are presented in Fig. \ref{f2}. 

Fig. \ref{f3} shows the histograms of the $v$ values  obtained for the $R$, $S$ and $K$ parameters in the case of the ensemble. The networks constituting the ensemble utilized lower $v$ values in majority, which may partially explain the lower performance of the network with the fixed value of $v$ equal to 2. 

\begin{table}[]
\begin{center}

        \caption{Estimation metrics calculated for the ensemble of neural networks. In this case, the moment order  parameter $v$ was trainable.}
        \label{t1}

\scalebox{0.75}{
\begin{tabular}{l c c c c}
\hline
    
      & ANN, trainable $v$  & ANN, $v$=2 & RSK & XU\\
                  \hline \hline
                  
    Mean absolute value of the relative bias of $u$ & 0.26 & 0.38 & 0.33 & 0.33  \\  \hline     
    
    Mean absolute value of the relative bias of $k$ & 0.89 & 0.93 & 0.62 & 0.35 \\  \hline     

    Mean normalized standard deviation of $u$ & 0.23 & 0.49 & 1.05 & 0.98 \\ \hline     
    
    Mean normalized standard deviation of $k$ & 0.04 & 0.04 & 0.61 &  0.90\\  \hline 
    
    Mean relative RMSE of $u$ & 0.37 & 0.66 & 1.11 & 1.04\\  \hline     
    
    Mean relative RMSE of $k$ & 0.89 & 0.93 & 0.93 & 0.99 \\  \hline

\end{tabular}} %($\pm$) \\ \cline{2-8}
\end{center}
\end{table}

\begin{table}[]
\begin{center}

        \caption{Estimation metrics calculated for the better performing single model. In this case, the moment order  parameter $v$ was trainable.}
        \label{t2}

\scalebox{0.75}{
\begin{tabular}{l c c c c}
\hline
    
     & ANN, trainable $v$  & ANN, $v$=2 & RSK & XU\\
                  \hline \hline
                  
    Mean absolute value of the relative bias of $u$ & 0.27 & 0.29 & 0.33 & 0.33 \\  \hline     
    
    Mean absolute value of the relative bias of $k$ & 0.76 & 0.80 & 0.62 & 0.35\\  \hline     

    Mean normalized standard deviation of $u$  & 0.27 & 0.36 & 1.05 & 0.98\\  \hline     
    
    Mean normalized standard deviation of $k$ & 0.16 & 0.04 & 0.61 & 0.90 \\  \hline 
    
    Mean relative RMSE of $u$ & 0.40 & 0.50 & 1.11 & 1.04 \\  \hline     
    
    Mean relative RMSE of $k$ & 0.79 & 0.81 & 0.93 & 0.99 \\  \hline

\end{tabular}} %($\pm$) \\ \cline{2-8}
\end{center}
\end{table}

\section{Conclusion}

In this work, we developed and evaluated a neural network for the estimation of the HK distribution parameters. Results demonstrated that the proposed method can be used to accurately estimate the parameters.

\section*{Conflicts of interest}

The authors do not have any conflicts of interest to disclosure.

\section*{Acknowledgement}

This work has not received funding. 

\bibliographystyle{IEEEtran}
\bibliography{mybibfile}

\end{document}